\documentclass[journal, hidelinks, 10pt]{IEEEtran}
\usepackage{amsmath,amsfonts,amssymb}
\usepackage{mathtools}
\usepackage{array}
\usepackage{textcomp}
\usepackage{stfloats}
\usepackage{url}
\usepackage{verbatim}
\usepackage{siunitx}
\usepackage{graphicx}
\usepackage{orcidlink}
\usepackage{cite}
\usepackage{xcolor}
\usepackage{fancyhdr}
\usepackage{bm}
\usepackage{cancel}
\usepackage{subcaption}

\usepackage[toc, acronym]{glossaries}
\glsdisablehyper
\def\BibTeX{{\rm B\kern-.05em{\sc i\kern-.025em b}\kern-.08em
    T\kern-.1667em\lower.7ex\hbox{E}\kern-.125emX}}
\usepackage{balance}
\usepackage[final]{microtype}
\usepackage{scalerel}
\usepackage{dsfont}

\newtheorem{remark}{Remark}

\DeclareMathOperator{\der}{\mathrm{d}\!}

\DeclareMathOperator{\expec}{\mathrm{E}}
\DeclareMathOperator{\trace}{\mathrm{tr}}

\newcommand{\prob}[2]{\ensuremath{\mathrm{Pr}\ifthenelse{\boolean{#2}}{\left[#1\right]}{[#1]}}}
\newcommand{\var}[3]{\ensuremath{\mathrm{var}_{#2}\ifthenelse{\boolean{#3}}{\left(#1\right)}{(#1)}}}    
\newcommand{\frob}{\ensuremath{\mathrm{F}}}   
\newcommand{\I}[3]{\ensuremath{\mathrm{I}\ifthenelse{\boolean{#3}}{\left(#1; #2\right)}{(#1; #2)}}}     
\newcommand{\h}[2]{\ensuremath{\mathrm{h}\ifthenelse{\boolean{#2}}{\left(#1\right)}{(#1)}}}             

\newcommand{\cov}[1]{\ensuremath{\mathbf{C}_{#1}}}          
\newcommand{\id}[1]{\ensuremath{\mathbf{I}_{#1}}}           
\newcommand{\bsf}[1]{\ensuremath{\boldsymbol{\mathsf{#1}}}} 
\newcommand{\herm}[1]{\ensuremath{#1^{\mathrm{H}}}}         
\newcommand{\trans}[1]{\ensuremath{#1^{\mathrm{T}}}}        
\newcommand{\placeholder}{\ensuremath{\boldsymbol{\cdot}}}

\newcommand{\reals}{\ensuremath{\mathbb{R}}}    
\newcommand{\complex}{\ensuremath{\mathbb{C}}}  
\newcommand{\imunit}{\ensuremath{\mathrm{j}}}   
\newcommand{\euler}{\ensuremath{\mathit{e}}}

\newcommand{\ie}{\textit{i.e. }}

\let\originalleft\left
\let\originalright\right
\renewcommand{\left}{\mathopen{}\mathclose\bgroup\originalleft}
\renewcommand{\right}{\aftergroup\egroup\originalright}

\newacronym{hsic}{HSIC}{Hilbert-Schmidt independence criterion}
\newacronym{jcf}{JCF}{joint characteristic function}
\newacronym{rkhs}{RKHS}{reproducing kernel Hilbert space}
\newacronym{chsic}{C-HSIC}{\textit{conditional} \acrshort{hsic}}
\newacronym{iid}{i.i.d.}{independent and identically distributed}

\newcommand{\changefont}{\color{blue}\fontsize{9}{9}\selectfont}

\let\oldmaketitle\maketitle
\renewcommand{\maketitle}{%
	\oldmaketitle
	\thispagestyle{fancy}
}

\begin{document}

    \title{Conditional Dependence via U-Statistics Pruning}

    \author{
        Ferran~de~Cabrera\,\orcidlink{0000-0001-6949-780X},
        Marc\,Vilà-Insa\,\orcidlink{0000-0002-7032-1411}, \IEEEmembership{Graduate Student Member, IEEE},
        and Jaume~Riba\,\orcidlink{0000-0002-5515-8169}, \IEEEmembership{Senior Member, IEEE}
        \thanks{This work was (partially) funded by project MAYTE (PID2022-136512OB-C21) by MICIU/AEI/10.13039/501100011033 and ERDF/EU, grant 2021 SGR 01033, and grant 2022 FI SDUR 00164 by Departament de Recerca i Universitats de la Generalitat de Catalunya.}\thanks{F. de Cabrera is an independent researcher (e-mail: ferrandecabrera@gmail.com). M. Vilà-Insa and J. Riba are with the Signal Processing and Communications Group (SPCOM), Universitat Politècnica de Catalunya (UPC), Barcelona, Spain (e-mail: marc.vila.insa@upc.edu; jaume.riba@upc.edu).}
        }

    \maketitle
    \begin{abstract}
        The problem of measuring conditional dependence between two random phenomena arises when a third one (a confounder) has a potential influence on the amount of information between them.
        A typical issue in this challenging problem is the inversion of ill-conditioned autocorrelation matrices.
        This paper presents a novel measure of conditional dependence based on the use of \textit{incomplete} unbiased statistics of degree two, which allows to re-interpret independence as uncorrelatedness on a finite-dimensional feature space.
        This formulation enables to prune data according to observations of the confounder itself, thus avoiding matrix inversions altogether.
        The proposed approach is articulated as an extension of the Hilbert-Schmidt independence criterion, which becomes expressible through kernels that operate on 4-tuples of data. 
    \end{abstract}
    
    
    \section{Introduction}\label{sec:intro}
    
        \IEEEPARstart{C}{onditional} dependence becomes relevant when a third phenomenon might explain, mediate, or confound the apparent relationship exhibited by an original random pair. 
        Its measurement is an important task in causal discovery and Bayesian network learning~\cite[Ch. 7]{zhang2011,zhang2008causal,peters2017elements}, which emerge in applied fields such as earth system sciences~\cite{runge2019inferring} or clinical data analysis~\cite{shen2020challenges}. 
        Due to its nature, empirically estimating conditional dependence is a complex problem, given that it also requires inferring conditional marginal distributions. 
        A known approach consists in measuring the Hilbert-Schmidt norm of the normalized conditional cross-covariance operator~\cite{Fukumizu2007}, grounded in the theory of \acrlong{rkhs}s (\acrshort{rkhs}). This kernel-based method is capable of measuring statistical dependence by gauging correlation of data implicitly mapped onto an infinite-dimensional space. However, it is also known to have numerical issues concerned with matrix inversions and regularization due to the usually low-rank structure of the involved correlation matrices~\cite{Fukumizu2007}.  This also leads to a fairly complex method that scales cubically with the observed data size, becoming prohibitive for large data sets.

        These issues have lately been the subject of interest of numerous lines of research. 
        A kernel-based conditional independence test is proposed in~\cite{zhang2011} by deriving the asymptotic distribution of the appropriate test statistic under the null hypothesis, which succeeds in avoiding matrix inversions, but whose complexity still increases cubically with the sample size.
        Other authors have proposed to approximate kernel functions by randomly selecting features~\cite{strobl2019approximate}, by providing an alternative review to the embedding process onto the \acrshort{rkhs}~\cite{park2020measure}, or by leveraging the interpretation that dependence produces correlation of distances~\cite{wang2015conditional}, \ie close pairs in one data set coincide with close pairs in the other~\cite{Gabor2007_dCor}. 
        Nevertheless, none of these methods addresses the aforementioned computational issues, since they all require matrix inversions. 
        An extension of the \acrlong{hsic} (\acrshort{hsic})~\cite{gretton2005}, a measure of unconditional independence, is introduced in~\cite{pfister2018kernel} with the objective of inferring causality, a similar endeavor to this letter, although focused on joint independence rather than conditional dependence. 

        This letter presents a procedure for measuring conditional dependence by statistically conditioning the observed data to a potential confounder in a simple way, bypassing matrix inversion problems.
        The basis for conditioning is to prune pairwise differences of data under the control of the confounder~\cite{vila2022}, stemming from distance-based methods~\cite{Gabor2007_dCor}.
        Simultaneously, the estimate is constructed as an alternative derivation of the kernel-based \acrshort{hsic}.
        While these two approaches are known to be equivalent~\cite{sheng2023distance}, we propose here an estimate that combines them, albeit for different tasks.
        To do so, we first provide a novel reinterpretation of the \acrshort{hsic} in Sec.~\ref{sec:marginal} by translating the problem of statistical dependence into one of correlation after mapping data onto finite-dimensional spaces based on steering vectors~\cite{cabrera2023}.
        Sec.~\ref{sec:u-stat} briefly reviews the theory of unbiased statistics (U-Statistics)~\cite{lee2019}, which is then employed in Sec.~\ref{sec:cond} to show that conditional dependence can be obtained by pruning U-Statistics.
        The obtained measure, named \acrfull{chsic}, is based on a classical signal processing structure but linked to kernel methods, which embraces the \acrshort{hsic} as a particular case when the U-Statistic is complete.
        
    \section{Marginal dependence as correlation}\label{sec:marginal}
    
        Let $\mathbf{d}:\reals\to\complex^{M}$ be a mapping based on windowed steering vectors.
        The $m$th element of $\mathbf{d}$ can be expressed as
        \begin{equation}
            [\mathbf{d}(\placeholder)]_m \triangleq \tfrac{1}{\sqrt[4]{M}} \mathrm{G}\bigl(\tfrac{m}{\sqrt{M}}\bigr) \euler^{\imunit\pi\frac{m}{\sqrt{M}}\placeholder}, \label{eq:mapping}
        \end{equation}
        where\footnote{$M$ is assumed even for mathematical convenience.} $m\in\{-M/2,\dots,M/2-1\}$ and $\mathrm{G}:\reals\to\reals$ is an even and absolutely integrable function with unit $L^2$-norm and maximum at $\mathrm{G}(0)$.
        Given two random variables $\mathsf{x}$ and $\mathsf{y}$, we define a pair of transformed random vectors\footnote{Without loss of generality, and for the sake of simplicity, we let $\bsf{u}$ and $\bsf{v}$ have the same dimensionality.} $\bsf{u}\triangleq\mathbf{d}(\mathsf{x})$ and $\bsf{v}\triangleq\mathbf{d}(\mathsf{y})$. Their cross-covariance matrix $\cov{\bsf{u},\bsf{v}}\in\complex^{M\times M}$ is defined as $\cov{\bsf{u},\bsf{v}} \triangleq \expec\bigl[\bsf{u}\herm{\bsf{v}}\bigr] - \expec[\bsf{u}]\herm{\expec[\bsf{v}]}$, where $\expec[\placeholder]$ denotes the expectation operator. 
        Then, the following implication holds~\cite{gretton2005}:
        \begin{equation}
            \lim_{M\to\infty} \lVert\cov{\bsf{u},\bsf{v}}\rVert_{\frob}^{2}=0 \quad\iff\quad \mathsf{x}, \mathsf{y} \text{ are independent},\label{eq:crit}
        \end{equation}
        where $\lVert\placeholder\rVert_{\frob}$ denotes the Frobenius norm.
        Terms $\expec[\mathbf{d}(\mathsf{x})]$ and $\expec[\mathbf{d}(\mathsf{y})]$ become dense samplings of the weighted (by $\mathrm{G}(\placeholder)$) marginal characteristic function of $\mathsf{x}$ and $\mathsf{y}$, respectively, in the limit of $M\to\infty$.
        Similarly, $\expec[\mathbf{d}(\mathsf{x})\herm{\mathbf{d}}(\mathsf{y})]$ becomes a dense sampling of the weighted \acrfull{jcf}.
        Therefore,~\eqref{eq:crit} is effectively checking the separability of the \acrshort{jcf} at every sample point (\ie the factorization of $\expec[\mathbf{d}(\mathsf{x})\herm{\mathbf{d}}(\mathsf{y})]$ as a product of expectations), which becomes equivalent to an uncorrelatedness property~\cite{cabrera-2018,cabrera2023}, akin to the \textit{distance covariance} in~\cite{Gabor2007_dCor}.
        
        Consider $L$ \acrshort{iid} samples of $\mathsf{x}$ and $\mathsf{y}$, $\{x(l),y(l)\}_{l=1,\dotsc,L}$, from which we obtain two transformed complex data matrices $\mathbf{U}\in\complex^{M\times L}$ and $\mathbf{V}\in\complex^{M\times L}$, defined as follows:
        \begin{subequations}
            \begin{align}
                \mathbf{U}&\triangleq[\mathbf{u}_1,\dotsc,\mathbf{u}_L]=[\mathbf{d}(x(1)),\dotsc,\mathbf{d}(x(L))] \\
                \mathbf{V}&\triangleq[\mathbf{v}_1,\dotsc,\mathbf{v}_L]=[\mathbf{d}(y(1)),\dotsc,\mathbf{d}(y(L))].
            \end{align}
        \end{subequations}
        The sample cross-covariance between the transformed vectors is given by
        \begin{align}
            \widehat{\mathbf{C}}_{\bsf{u},\bsf{v}} & \triangleq \tfrac{1}{L-1}\sum_{l=1}^L\Biggl(\mathbf{u}_l-\tfrac{1}{L}\sum_{i=1}^L \mathbf{u}_i\Biggr)\herm{\Biggl(\mathbf{v}_l-\tfrac{1}{L}\sum_{j=1}^L \mathbf{v}_j\Biggr)} \nonumber \\
            & = \tfrac{1}{L-1}\mathbf{U}\mathbf{P}\herm{\mathbf{V}}, \quad \label{eq:sample_cov}
        \end{align}
        where $\mathbf{P} \triangleq \id{}-\tfrac{1}{L}\mathbf{1}\trans{\mathbf{1}}\in\reals^{L\times L}$ is the \textit{data centering matrix}, a well-known projection matrix in the signal processing and kernel methods literature~\cite[Appx. B.7]{ramirez2023coherence}.
        Using~\eqref{eq:crit}, a marginal dependence measure is given by 
        \begin{align} \label{eq:trace}
            \lVert\widehat{\mathbf{C}}_{\bsf{u},\bsf{v}}\rVert_{\frob}^2=\trace(\herm{\widehat{\mathbf{C}}_{\bsf{u},\bsf{v}}}\widehat{\mathbf{C}}_{\bsf{u},\bsf{v}})&=\trace\left(\tfrac{1}{(L-1)^{2}}\mathbf{V}\herm{\mathbf{P}}\herm{\mathbf{U}}\mathbf{U}\mathbf{P}\herm{\mathbf{V}}\right) \nonumber \\
            & =\tfrac{1}{(L-1)^{2}}\trace(\mathbf{P}\herm{\mathbf{U}}\mathbf{U}\mathbf{P}\herm{\mathbf{V}}\mathbf{V}),
        \end{align}
        where the circularity of the trace operator has been used.
        To see the link of~\eqref{eq:trace} with the \acrshort{hsic}, let us examine the limit for $M\to\infty$ of the kernel matrices $\mathbf{K}\triangleq\lim_{M\rightarrow\infty}\herm{\mathbf{U}} \mathbf{U}\in\reals^{L\times L}$ and $\mathbf{Q}\triangleq\lim_{M\rightarrow\infty}\herm{\mathbf{V}} \mathbf{V}\in\reals^{L\times L}$.
        The elements of $\mathbf{K}$ (and analogously those of $\mathbf{Q}$) are the following:
        \begin{align}
            [\mathbf{K}]_{l,l'} &= \smashoperator{\lim_{M\to\infty}} \herm{\mathbf{d}}(x(l))\mathbf{d}(x(l')) \nonumber \\
            &= \smashoperator[l]{\lim_{M\to\infty}} \smashoperator{\sum_{m=-\frac{M}{2}}^{\frac{M}{2}-1}} \tfrac{1}{\sqrt{M}}\mathrm{G}^{2}\bigl(\tfrac{m}{\sqrt{M}}\bigr) \euler^{-\imunit2\pi(x(l)-x(l'))\frac{m}{\sqrt{M}}} \label{eq:kernel} \\
            &= \int_{\mathrlap{-\infty}}^{\mathrlap{\infty}} \mathrm{G}^{2} (f) \euler^{-\imunit2\pi(x(l)-x(l'))f}
            \der f \triangleq \kappa\bigl(x(l)-x(l')\bigr) \nonumber,
        \end{align}
        where $\kappa(\placeholder)$ is the kernel function that results from the Fourier transform\footnote{
            From the properties imposed on $\mathrm{G}^{2}(f)$, $\kappa(\placeholder)$ becomes naturally an autocorrelation translation-invariant kernel, such that $\kappa(0)=1\geq\lvert\kappa(s)\rvert$ and $\kappa(\infty)=\kappa(-\infty)=0$. The reader is referred to~\cite[Sec.~1.4]{rudin1990} where these ideas emerge in the light of Bochner's theorem.
        } of $\mathrm{G}^{2}(f)$, and the integral is the limit of the Darboux sum in the second line.
        As a result, the entries of $\mathbf{K}$ and $\mathbf{Q}$ are just the evaluation of $\kappa(\placeholder)$ at the difference between two data samples.
        Then, given \eqref{eq:trace} and taking the limit $M\to\infty$, we obtain the \acrshort{hsic}~\cite[Sec. 3.1]{gretton2005}:
        \begin{equation}
            \text{\acrshort{hsic}}\left(\mathbf{x};\mathbf{y}\right)\triangleq \tfrac{1}{(L-1)^{2}}\trace(\mathbf{P}\mathbf{K}\mathbf{P}\mathbf{Q}) = \smashoperator[l]{\lim_{M\to\infty}} \lVert\widehat{\mathbf{C}}_{\bsf{u},\bsf{v}}\rVert_{\frob}^2,\label{eq:hsic}
        \end{equation}
        with $\mathbf{x}=\trans{[x(1),\dotsc,x(L)]}$ and $\mathbf{y}=\trans{[y(1),\dotsc,y(L)]}$.
        
        In summary, the alternative formulation exposed above shows that the \acrshort{hsic} results from measuring correlation in a finite-dimensional space based on steering vectors, where the kernel formulation arises in a second stage once the dimensionality grows without bound. The interplay between kernels and cross-covariance matrices of mapped data will be used later on to condition the kernel matrices by pruning data.
        Beforehand, we will reformulate the estimation of cross-covariance matrices leveraging incomplete U-statistics, which will serve as the basis for data pruning.
    
    \section{Sample covariance matrix as an incomplete U-Statistic}\label{sec:u-stat}

        Consider a list containing all the unique $K_{\max}\triangleq L(L-1)/2$ pairs that can be constructed from $L$ different samples of a random variable.
        This list is ordered arbitrarily such that a single index $k$ identifies both elements of a pair.
        Let $\mathrm{f}_1(k)$ and $\mathrm{f}_2(k)$ be two functions that return the corresponding indices of each term of the $k$th pair.
        Given $K\leq K_{\max}$ indices, we construct the pairwise differences
        \begin{equation} \label{eq:sample_pairs}
            \mathring{\mathbf{u}}_k \triangleq \tfrac{1}{\sqrt{2}}\left(\mathbf{u}_{\mathrm{f}_1(k)}-\mathbf{u}_{\mathrm{f}_2(k)}\right),\quad \mathring{\mathbf{v}}_k \triangleq \tfrac{1}{\sqrt{2}}\left(\mathbf{v}_{\mathrm{f}_1(k)}-\mathbf{v}_{\mathrm{f}_2(k)}\right),
        \end{equation}
        for $k\in\{1,\dots,K\}$, corresponding to the samples of two virtual sources $\mathring{\bsf{u}}$ and $\mathring{\bsf{v}}$.
        It is worth noting that these are zero-mean variables for any pair with $\mathrm{f}_1(k)\neq\mathrm{f}_2(k)$, since $\bsf{u}$ and $\bsf{v}$ are \acrshort{iid}.
        Their sample cross-covariance matrix $\widehat{\mathbf{C}}_{\mathring{\bsf{u}},\mathring{\bsf{v}}}\in\complex^{M\times M}$ is then the following:
        \begin{equation}
            \widehat{\mathbf{C}}_{\mathring{\bsf{u}},\mathring{\bsf{v}}}=\tfrac{1}{K}\mathring{\mathbf{U}}\herm{\mathring{\mathbf{V}}},\label{eq:sample_cov_2}
        \end{equation}
        where $\mathring{\mathbf{U}}\triangleq[\mathring{\mathbf{u}}_{1},\dots,\mathring{\mathbf{u}}_{K}]\in\complex^{M\times K}$ and $\mathring{\mathbf{V}}\triangleq[\mathring{\mathbf{v}}_{1},\dots,\mathring{\mathbf{v}}_{K}]\in\complex^{M\times K}$.
        Note that, thanks to the constant factor in~\eqref{eq:sample_pairs}, $\mathring{\bsf{u}}$ and $\mathring{\bsf{v}}$ have the same average covariance matrix as $\bsf{u}$ and $\bsf{v}$, respectively. 
        In contrast to~\eqref{eq:sample_cov}, $\mathbf{P}$ is missing in~\eqref{eq:sample_cov_2} as a result of constructing zero-mean virtual data in~\eqref{eq:sample_pairs}, which will lead to a cleaner implementation and fewer matrix operations. 
        
        Equation~\eqref{eq:sample_cov_2} is in fact an instance of a U-Statistic.
        For $K=K_{\text{max}}$, \ie when all data pairs are taken, we obtain the equality $\widehat{\mathbf{C}}_{\mathring{\bsf{u}},\mathring{\bsf{v}}}=\widehat{\mathbf{C}}_{\bsf{u},\bsf{v}}$ from U-Statistics theory~\cite{lee2019}. 
        In contrast, for $K< K_{\text{max}}$, \ie when~\eqref{eq:sample_cov_2} is an incomplete U-Statistic, although $\widehat{\mathbf{C}}_{\mathring{\bsf{u}},\mathring{\bsf{v}}}$ remains unbiased, its estimation variance increases due to the pruning of data. 
        Nevertheless, $\widehat{\mathbf{C}}_{\mathring{\bsf{u}},\mathring{\bsf{v}}}$ is still a consistent estimate of $\cov{\bsf{u},\bsf{v}}$ provided that $K\rightarrow\infty$ as $L\rightarrow\infty$.

        \begin{remark}[Robustness to pruning] \label{remark:robust}
            A notable property of incomplete U-statistics is that their robustness against data pruning increases the larger $L$ is~\cite{vila2022}.
            To briefly illustrate this property, consider taking only the $K=\left\lfloor L/2\right\rfloor$ data pairs with no indices in common.
            The number of remaining, unused, data pairs is equal to $L(L-1)/2-\lfloor L/2\rfloor$, which increases with $O(L^{2})$.
            To only use the \acrshort{iid}, unique, terms is effectively equivalent to computing $\widehat{\mathbf{C}}_{\bsf{u},\bsf{v}}$ with half of the available samples~\cite{chen-2019}. 
            It is then safe to assume that their contribution to the overall accuracy of the sample covariance is higher than those with repeated indices. 
            Therefore, the larger $L$ is, the higher the amount of pairs that can be pruned for some specified degradation in the estimation accuracy of the resulting sample covariance. 
            The implication is that $K$ in the incomplete U-Statistic can be designed to grow with $O(L)$ instead of $O(L^{2})$, which provides considerable flexibility for pruning, eases the overall computational complexity, and will be used for choosing the number of pairs in the next section.
        \end{remark}

        Once we have determined the incomplete U-statistic formulation of the cross-covariance matrix, we are now in a position to establish the underlying rule of $\mathrm{f}_1(\placeholder)$ and $\mathrm{f}_2(\placeholder)$ by looking at sample pairwise differences of the confounder.
        Afterwards, we will be employing the pruned cross-covariance matrix to return to kernel methods in the same way as portrayed in~\eqref{eq:hsic}.

    \section{Conditional dependence via U-Statistics}\label{sec:cond}
        
        The conditional cross-covariance matrix between $\bsf{u}$ and $\bsf{v}$ is defined as:
        \begin{equation}
            \cov{\bsf{u},\bsf{v}\vert\mathsf{z}}
            \triangleq\int_{\reals}\cov{\bsf{u},\bsf{v}\vert\mathsf{z}=z}\der F_{\mathsf{z}}(z)
            =\int_{\reals}\cov{\mathring{\bsf{u}},\mathring{\bsf{v}}\vert\mathsf{z}=z}\der F_{\mathsf{z}}(z),\label{eq:cond}
        \end{equation}
        where $\cov{\mathring{\bsf{u}},\mathring{\bsf{v}}\vert\mathsf{z}=z}=\cov{\bsf{u},\bsf{v}\vert\mathsf{z}=z}$ (given~\eqref{eq:sample_pairs}) is the cross-covariance matrix conditioned to a specific value $z$ of confounder $\mathsf{z}$, and $F_{\mathsf{z}}(z)$ is its cumulative distribution.
        With the goal of deriving an estimator of~\eqref{eq:cond}, we define the virtual random variable $\mathring{\mathsf{z}}\triangleq\mathsf{z}_1-\mathsf{z}_2$, where $\mathsf{z}_1$ and $\mathsf{z}_2$ are mutually independent and distributed as $\mathsf{z}$.
        Integrating over all values of $\mathsf{z}$ is equivalent to doing so over the events in which $\mathsf{z}_1$ and $\mathsf{z}_2$ take the same value, \ie $\mathring{\mathsf{z}}=0$.
        Therefore, the expectation in~\eqref{eq:cond} can be alternatively expressed as
        \begin{align}    \label{eq:cond-1-1}                        
            \cov{\bsf{u},\bsf{v}\vert\mathsf{z}} & = \iint_{\reals^{2}} \cov{\mathring{\bsf{u}},\mathring{\bsf{v}}\vert\mathring{\mathsf{z}}=0}\der F_{\mathsf{z}}(z_{1})\der F_{\mathsf{z}}(z_{2}) \nonumber \\
            & = \cov{\mathring{\bsf{u}},\mathring{\bsf{v}}\vert\mathring{\mathsf{z}}=0} \iint_{\reals^{2}} \der F_{\mathsf{z}}(z_{1})\der F_{\mathsf{z}}(z_{2}) = \cov{\mathring{\bsf{u}},\mathring{\bsf{v}}\vert\mathring{\mathsf{z}}=0}, 
        \end{align}
        where $\iint_{\reals^{2}} \der F_{\mathsf{z}}(z_{1})\der F_{\mathsf{z}}(z_{2})=1$, and $\cov{\mathring{\bsf{u}},\mathring{\bsf{v}}\vert\mathring{\mathsf{z}}=0}$ does not depend on the specific values of $z_1$ and $z_2$ but rather on them being equal.
        In consequence, conditioning with respect to $\mathsf{z}$ is equal to conditioning with respect to $\mathring{\mathsf{z}}=0$.
        Therefore,~\eqref{eq:cond-1-1} suggests to let the pruning of the incomplete U-Statistics in~\eqref{eq:sample_cov_2} be handled by the potential confounder.
        However, since $\mathring{\mathsf{z}}=0$ is an event of zero probability for continuous random variables, data control should be based on merely small values of $\lvert\mathring{z}\rvert$~\cite{vila2022}. 
        
        With the intention of choosing data pairs to prune according to $\lvert\mathring{z}\rvert$, we define the samples of $\mathring{\mathsf{z}}$ as follows:
        \begin{equation}
            \mathring{z}_{l,l'}\triangleq z(l)-z(l'),\qquad l\neq l',\label{eq:virz}
        \end{equation}
        where $z(l)$ and $z(l')$ are \acrshort{iid} samples drawn from $\mathsf{z}$.
        Then, we let the sorting of $\lvert\mathring{z}_{l,l'}\rvert$ (in ascending order) be the one that determines the ordering of the index pairs provided by $\mathrm{f}_1(\placeholder)$ and $\mathrm{f}_2(\placeholder)$ in Sec.~\ref{sec:u-stat}. 
        Moreover, in view of \textit{Remark 1}, the amount of pairs $K$ is set to grow as $O(L)$ with
        \begin{equation}
            K=\bigl\lfloor\tfrac{L\alpha}{2}\bigr\rfloor,\label{eq:propo}
        \end{equation}
        being $1\leq\alpha\ll(L-1)$ a tuning hyper-parameter. While $\alpha=1$ ensures that only very small values of $\lvert\mathring{z}_{l,l'}\rvert$ are considered, the U-Statistic becomes complete and there is no conditioning at all for $\alpha=L-1$, thus yielding the \acrshort{hsic} as a particular case as in~\eqref{eq:hsic}. 
        Then a natural trade-off emerges: low values of $\alpha$ are desirable to provide strong conditioning of data, but may also lead to excessive pruning with low statistical accuracy in~\eqref{eq:sample_cov_2}.
        Remarkably, the selection of $\alpha$ becomes a minor issue provided that $L$ is sufficiently large, as shown in~\cite{vila2022} under the correlation measure framework between a pair of vectors. This will be further discussed in Sec.~\ref{sec:numerical} with a numerical example.
        
        \subsection{Conditional HSIC}
        Now that we have determined the sorting and pruning of data pairs according to the confounder $\mathsf{z}$, let us write a conditional dependence measure as the Frobenius norm of~\eqref{eq:sample_cov_2}:
            \begin{equation} \label{eq:incomplete_Frob}
                 \trace\bigl(\herm{\widehat{\mathbf{C}}}_{\mathring{\bsf{u}},\mathring{\bsf{v}}\vert\mathsf{z}}\widehat{\mathbf{C}}_{\mathring{\bsf{u}},\mathring{\bsf{v}}\vert\mathsf{z}}\bigr) = \tfrac{1}{K^{2}}\trace\bigl( \herm{\mathring{\mathbf{U}}}\mathring{\mathbf{U}}\herm{\mathring{\mathbf{V}}}\mathring{\mathbf{V}}\bigr),
            \end{equation}
        where the circularity of the trace has been used. 
        To link the previous expression with kernel-based methods, let us rewrite the zero-mean virtual data matrix $\mathring{\mathbf{U}}$ as follows:
        \begin{equation}
            \mathring{\mathbf{U}}=\tfrac{1}{\sqrt{2}}(\mathbf{U}_{1}-\mathbf{U}_{2}),\quad
            \mathbf{U}_{a}\triangleq[\mathbf{u}_{\mathrm{f}_a(1)},\dots,\mathbf{u}_{\mathrm{f}_a(K)}], \label{eq:virtuxy}
        \end{equation}
        for $a=\{1,2\}$.
        The same is done for $\mathring{\mathbf{V}}$.
        Accordingly,~\eqref{eq:incomplete_Frob} is then rewritten as
        \begin{align} \label{eq:incomplete_Frob_2}
                 &\trace\bigl(\herm{\widehat{\mathbf{C}}}_{\mathring{\bsf{u}},\mathring{\bsf{v}}\vert\mathsf{z}}\widehat{\mathbf{C}}_{\mathring{\bsf{u}},\mathring{\bsf{v}}\vert\mathsf{z}}\bigr)  =\\
                 &\quad\tfrac{1}{4K^{2}}\trace\bigl(\herm{(\mathbf{U}_{1}-\mathbf{U}_{2})}(\mathbf{U}_{1}-\mathbf{U}_{2})\herm{(\mathbf{V}_{1}-\mathbf{V}_{2})}(\mathbf{V}_{1}-\mathbf{V}_{2})\bigr).\nonumber
            \end{align}
        Taking the limit of $M\rightarrow\infty$, kernel matrices are obtained from the products among $\mathbf{U}$ and $\mathbf{V}$ in~\eqref{eq:incomplete_Frob_2}:
        \begin{equation}
            \mathbf{K}_{a,a'}\triangleq\lim_{M\rightarrow\infty}\herm{\mathbf{U}_{a}}\mathbf{U}_{a'}, \qquad 
            \mathbf{Q}_{a,a'}\triangleq\lim_{M\rightarrow\infty}\herm{\mathbf{V}_{a}}\mathbf{V}_{a'},
        \end{equation}
        whose elements are
        \begin{subequations}
            \begin{align}
                [\mathbf{K}_{a,a'}]_{k,k'} & =\kappa \bigl(x(\mathrm{f}_a(k))-x(\mathrm{f}_{a'}(k'))\bigr), \\
                [\mathbf{Q}_{a,a'}]_{k,k'} & =\kappa \bigl(y(\mathrm{f}_{a}(k))-y(\mathrm{f}_{a'}(k'))\bigr),
            \end{align}
        \end{subequations}
        where $\kappa(\placeholder)$ is the kernel function as in~\eqref{eq:kernel}.
        Finally, the resulting \acrshort{chsic} can be expressed as follows:
        \begin{equation}
            \text{C-HSIC}_{\alpha}(\mathbf{x};\mathbf{y})\triangleq\tfrac{1}{4K^{2}}\trace\bigl(\Breve{\mathbf{K}}\Breve{\mathbf{Q}}\bigr)\label{eq:c-hsic}
        \end{equation}
        with $\Breve{\mathbf{K}} \triangleq\mathbf{K}_{1,1}+\mathbf{K}_{2,2}-\mathbf{K}_{1,2}-\herm{\mathbf{K}_{1,2}}$ and $\Breve{\mathbf{Q}} \triangleq\mathbf{Q}_{1,1}+\mathbf{Q}_{2,2}-\mathbf{Q}_{1,2}-\herm{\mathbf{Q}_{1,2}}$.
        Note that, as a result of the U-Statistics implementation, each entry of the new kernel-based matrices involves four data samples of the same source ($4$-tuples), in contrast to only the pairs typically involved in classical kernel methods. 
        This fact, along with the lack of $\mathbf{P}$, are the main distinctive features of the \acrshort{chsic}~\eqref{eq:c-hsic} vs. \acrshort{hsic}~\eqref{eq:hsic}. In contrast to other conditional dependence measures,~\eqref{eq:c-hsic} does not require any matrix inversion. 
        Since $K$ is set to grow linearly with $L$ in~\eqref{eq:propo}, the computational complexity is $O(L^2)$.
    
    \section{Numerical illustrations} \label{sec:numerical}
    
        To test the proposed method, we aim at generating uncorrelated data with a controlled amount of conditional dependence. 
        Two scenarios are studied, $\mathcal{M}^{+}$ and $\mathcal{M}^{-}$, defined as follows:
        \begin{equation} \label{eq:models}
            \mathcal{M}^{+}:\begin{cases}
                \mathsf{x}=\sqrt{\gamma}\mathsf{a}\mathsf{p}+\mathsf{v}\\
                \mathsf{y}=\sqrt{\gamma}\mathsf{a}\mathsf{q}+\mathsf{w}\\
                \mathsf{z}=\mathsf{a}
            \end{cases}\hspace{-.5em},\hspace{.5em}
            \mathcal{M}^{-}:\begin{cases}
                \mathsf{x}=\sqrt{\gamma}\mathsf{b}\mathsf{p}+\mathsf{v}\\
                \mathsf{y}=\sqrt{\gamma}\mathsf{c}\mathsf{q}+\mathsf{w}\\
                \mathsf{z}=\mathsf{b}-\mathsf{c}
            \end{cases}\hspace{-.5em}.
        \end{equation}
        The internal \acrshort{iid} random variables are distributed as $\mathsf{a},\mathsf{b},\mathsf{c}\sim\mathcal{U}(0,\sqrt{3\text{}})$ (uniform), $\mathsf{v},\mathsf{w}\sim\mathcal{N}(0,1)$ (normal), and $\mathsf{p},\mathsf{q}\sim\mathrm{Bern}_{1/2}\{-1,1\}$ (equiprobable Bernoulli).
        In $\mathcal{M}^{+}$, the pair $\{\mathsf{x},\mathsf{y}\}$ are dependent, \ie their mutual information is greater than zero, but they are conditionally independent, since knowing $\mathsf{z}$ implies that $\mathsf{x}$ and $\mathsf{y}$ become solely driven by independent phenomena ($\mathsf{v}$ and $\mathsf{w}$).  
        By contrast, $\mathsf{x}$ and $\mathsf{y}$ are marginally independent in $\mathcal{M}^{-}$, but become conditionally dependent, since knowledge of $\mathsf{z}$ correlates the possible joint values of $\mathsf{b}$ and $\mathsf{c}$. 
        Parameter $\gamma$ is the signal-to-noise ratio associated with the measurements and controls the total amount of unconditional dependence in $\mathcal{M}^{+}$ and conditional dependence in $\mathcal{M}^{-}$.
        In both models, $\mathsf{x}$, $\mathsf{y}$ and $\mathsf{z}$ are mutually uncorrelated due to the multiplicative effect of mutually independent variables $\mathsf{p}$ and $\mathsf{q}$, so correlation measures are unable to discover any data association.
        Similar ideas on modeling conditional dependencies can be found in~\cite{Poczos-2012}, which are inspired by co-information~\cite{Bell2003,Ghassami2017}.

        The universal Gaussian kernel is used as the kernel function of choice, which yields $\kappa(s)=\exp(-(\frac{s}{\hat{\sigma}L^{-1/5}})^{2})$ being $\hat{\sigma}$ the sample standard deviation. 
        We use this expression because of its association with kernel density estimation, known to be related to kernel methods when estimating certain dependence measures~\cite[Ch. 2]{principe_2010}. 
        This connection also justifies the adoption of the power law $O(L^{-1/5})$~\cite[Ch. 3]{silverman-86},~\cite[Appx.~D]{cabrera2023}.

        \begin{figure}[tp]
            \begin{subfigure}{\columnwidth}
                \includegraphics[width=\columnwidth]{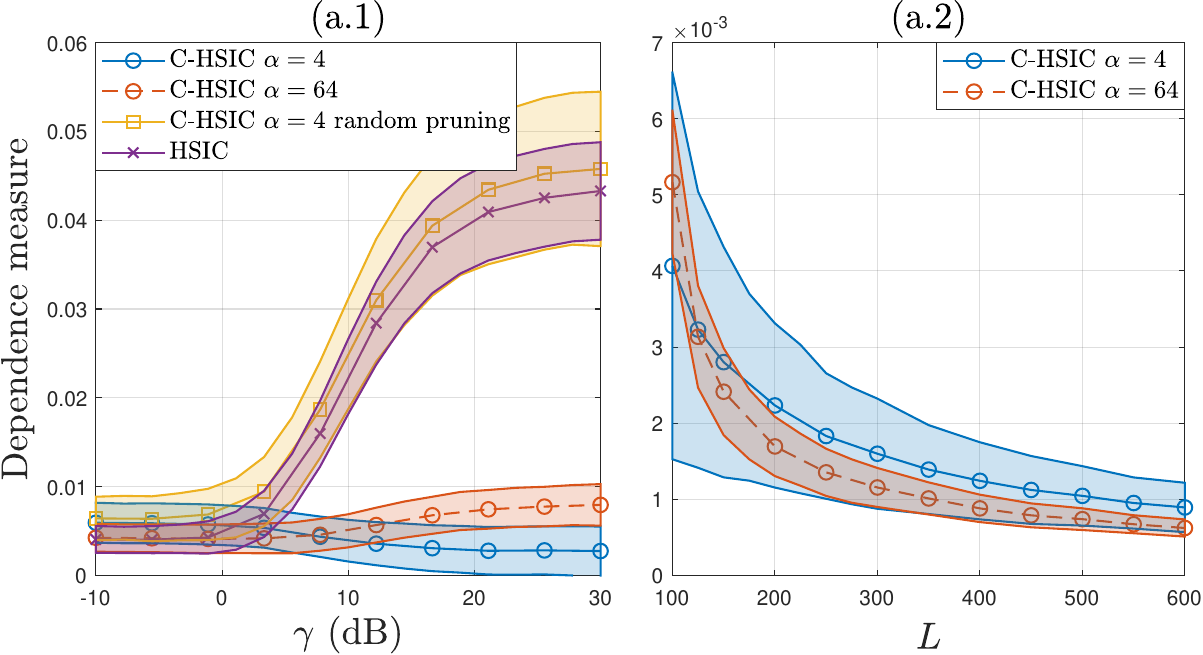}
                \caption{Model $\mathcal{M}^{+}$.}
            \end{subfigure}
            \begin{subfigure}{\columnwidth}
                \includegraphics[width=\columnwidth]{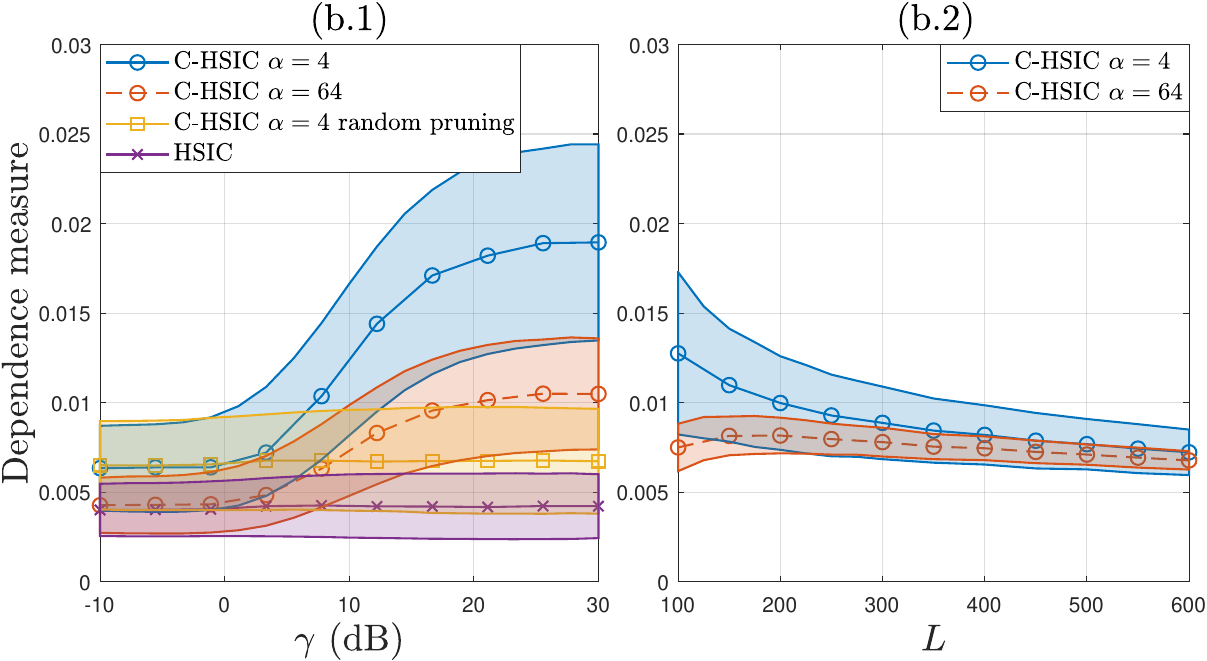}
                \caption{Model $\mathcal{M}^{-}$.}
            \end{subfigure}
            \caption{Dependence measures as a function of $\gamma$ (with $L=100$) and $L$ (with $\gamma=10$dB). \acrshort{chsic} measures conditional dependence while \acrshort{hsic} measures unconditional dependence. Markers denote the empirical average and bands indicate the standard deviation.}
            \label{fig:fig1}
        \end{figure}
        
        Fig.~\ref{fig:fig1} shows measures of dependence for both models with two $\alpha$ values from~\eqref{eq:propo}: a moderate $\alpha=4$ and a sixteenfold increase $\alpha=64$. This yields $4\%$ and $64.5\%$, respectively, of the total data pairs for $L=100$, or $0.7\%$ and $10.7\%$ for $L=600$ (and even less for higher $L$). 
        For $\mathcal{M}^{+}$ (Fig.~\ref{fig:fig1}-(a)), \acrshort{chsic} correctly depicts conditional independence for an increasing value of $\gamma$, while \acrshort{hsic} confirms that marginal dependence is high as $\gamma$ increases. 
        Conversely, Fig.~\ref{fig:fig1}-(b) exhibits the capability of \acrshort{chsic} to discover conditional dependence for moderate and high $\gamma$ values in $\mathcal{M}^{-}$, while marginal dependence is confirmed to be small by \acrshort{hsic} at any value of $\gamma$. 

        To get further insights on the proposed ideas, Fig.~\ref{fig:fig1}-(a.1) and Fig.~\ref{fig:fig1}-(b.1) also show the performance of \acrshort{chsic} when pruning data pairs is performed randomly in the U-statistic, \ie the pair given by $\mathrm{f}_1(\placeholder)$ and $\mathrm{f}_2(\placeholder)$ is not controlled by the confounder. 
        As a result of not being properly conditioned, \acrshort{chsic} produces a measure of marginal dependence similar to \acrshort{hsic}, but with increased variance due to the pruning itself. 
        It is also worth noting that random and orderly pruning perform similarly for low values of $\gamma$, corresponding to the regime where the confounder has no influence on the pair $\{\mathsf{x},\mathsf{y}\}$, and start to deviate as $\gamma$ increases. 
        Additionally, it can be seen that the expected value deteriorates for $\alpha=64$, increasing in $\mathcal{M}^{+}$ and decreasing in $\mathcal{M}^{-}$. 
        This is a consequence of a mild pruning for $L=100$ (where $\alpha=64$ is intentionally chosen to illustrate this effect), worsening the conditioning process and starting to behave as an unconditioned measure. 

        Finally, Fig.~\ref{fig:fig1}-(a.2) and Fig.~\ref{fig:fig1}-(b.2) illustrate the behavior of \acrshort{chsic} for multiple values of $L$ at $\gamma=10$dB. 
        Model $\mathcal{M}^{+}$, which is conditionally independent, shows that \acrshort{chsic} correctly approaches zero as $L$ increases. 
        Conversely, it tends to some nonzero value in $\mathcal{M}^{-}$. 
        As mentioned above, $\alpha=64$ is using too many data pairs for small $L$ values and provides a degraded empirical average. 
        In both models, $\alpha=4$ shows an increased variance with respect to $\alpha=64$ due to a more thorough pruning. 
        However, this is reduced for sufficiently large data sizes due to \textit{Remark \ref{remark:robust}}. 
        Similarly, the gap between expected values for $\alpha=4$ and $\alpha=64$ is reduced as $L$ increases, albeit the former has the advantage of a lower computational complexity.
        These results show that $\alpha$ is a noncritical hyper-parameter, provided it is reasonably small. 
    
    \section{Conclusions}

        This work has proposed a proof of concept for a modification on the classical \acrshort{hsic} under the conditional dependence framework, named \acrshort{chsic}. 
        Its interpretation as a correlation measure on a finite but high-dimensional space allows an insightful connection with kernels by letting the dimension grow without bound.
        Moreover, it opens the possibility of leveraging U-Statistics for this task.
        Thanks to this formulation, we can provide a novel approach for performing statistical conditioning by pruning data pairs based on the pairwise differences of a potential confounder. 
        Furthermore, the proposed measure of conditional dependence does not require matrix inversions, which has the advantage of reduced computational complexity and the avoidance of addressing ill-conditioned matrices.
        Numerical illustrations have shown that \acrshort{chsic} is capable of measuring both conditional dependence and independence in two different scenarios. 
        Further research should study the potential of the proposed method with richer data sets, as well as provide a thorough comparison with other methods for measuring conditional dependence.

\end{document}